\documentclass[runningheads]{llncs}
\usepackage{graphicx}
\usepackage{amsmath}
\usepackage{amssymb}
\usepackage{booktabs}
\usepackage{float}
\usepackage{mathtools}
\usepackage{stfloats}
\usepackage{tabularx}
\usepackage{adjustbox}
\usepackage{colortbl}
\usepackage{multirow}
\usepackage[accsupp]{axessibility}

\usepackage[pagebackref,breaklinks,colorlinks]{hyperref}

\usepackage[capitalize]{cleveref}
\crefname{section}{Sec.}{Secs.}
\Crefname{section}{Section}{Sections}
\Crefname{table}{Table}{Tables}
\crefname{table}{Tab.}{Tabs.}

\newcommand\ie {{i.e., }}

\begin{document}
\title{Adversarial Consistency for Single Domain Generalization in Medical Image Segmentation}
%
%
\author{Yanwu Xu\inst{1} \and
Shaoan Xie\inst{3} \and
Maxwell Reynolds\inst{1} \and
Matthew Ragoza\inst{1} \and 
Mingming Gong*\inst{2} \and
Kayhan Batmanghelich*\inst{1} (*Equal Contribution)}
\authorrunning{Yanwu. Author et al.}
%
\institute{Department of Biomedical Informatics, University of Pittsburgh \and
School of Mathematics and Statistics, The University of Melbourne\\
\and
Department of Philosophy, Carnegie Mellon University}
\maketitle              
\vspace{-8mm}
\begin{abstract}

An organ segmentation method that can generalize to unseen contrasts and scanner settings can significantly reduce the need for retraining of deep learning models. Domain Generalization (DG) aims to achieve this goal. However, most DG methods for segmentation require training data from multiple domains during training. We propose a novel adversarial domain generalization method for organ segmentation trained on data from a \emph{single} domain. We synthesize the new domains via learning an adversarial domain synthesizer (ADS) and presume that the synthetic domains cover a large enough area of plausible distributions so that unseen domains can be interpolated from synthetic domains. We propose a mutual information regularizer to enforce the semantic consistency between images from the synthetic domains, which can be estimated by patch-level contrastive learning. We evaluate our method for various organ segmentation for unseen modalities, scanning protocols, and scanner sites.  
\vspace{-3mm}
\end{abstract}
\section{Introduction}
Deep Learning-based methods for the segmentation of medical images hold state-of-the-art performance across various organs and anatomies~\cite{unet,vnet,bda}. The independent and identically distributed ($\textit{i.i.d.}$) is the underlying assumption of most of those methods.However, the difference in the image acquisition, such as scanning protocol and image modality, introduces domain shifts, rendering the assumption impractical.Domain Adaptation~\cite{dann,dsn,ada,cycada} (DA) and Multi-source Domain generalization~\cite{dgm} (MDG) aim to alleviate the domain shift issue.However, those approaches are not data-efficient as they either require access to test distribution (\ie DA) or need multiple labeled source domains during training (\ie MDG). In this paper, we focus on \emph{single-source} domain generalization (SDG), aiming to train a generalizable deep model on only one source domain.

The $SDG$ does not require access to the test distribution or labeled data from multiple sources during the training. As a result, it reduces annotation costs and avoids repetitive adaptation for each new domain. 
In the literature, various SDG methods have been proposed that are based on augmentation of input image~\cite{gin,randconv} and meta learning~\cite{sdg_adversarial2}. The meta-learning techniques tend to be extremely slow during inference time. The augmentation methods synthesize new images using random initialization of convolution filter~\cite{gin,randconv}. However, those methods cannot avoid over-fitting to a regular pattern of synthetic data. Thus, we propose synthesizing the new domains via learning an adversarial framework.

We propose synthesizing the new domains via learning an adversarial domain synthesizer (ADS). The intuition is that the synthetic domains cover a large enough area of plausible distributions so that unseen domains can be interpolated from synthetic domains. 
Specifically, we design the synthesizer with a random style module, enabling ADS to synthesize random textures during adversarial training. 
Without a constraint, adversarial training may change the image semantics, making the synthetic domains irrelevant.
To remedy this problem, we propose to keep the underlying semantic information between the source image and the synthetic image via a mutual information regularizer. As estimating mutual information is hard for high dimensional data, we utilize the patch-level contrastive loss~\cite{info} as a surrogate to maximize the mutual information between the original and synthesized images.

The main contributions of this work can be summarized as follows: 1) We propose an adversarial framework for single domain generalization of medical image segmentation.
2) We redesign the network structure of synthesizing new domains. 3) To constrain the adversarial training, we propose a regularization method for synthetic images. To evaluate our model, we conduct experiments of single domain generalization of medical image segmentation on cross-modality image segmentation (CT $\rightarrow$ MRI), -imaging protocol, and -organizations.

\vspace{-3mm}
\section{Related Works}
\vspace{-3mm}
\paragraph{Unsupervised Domain Adaptation and Domain Generalization}
Unsupervised Domain Adaptation (UDA) is a proposed strategy for addressing domain shift between the training data and testing data of deployed applications. In the UDA framework, the model has access to labelled data from a source domain and unlabelled data from a target domain during training. Prior work on UDA can be classified as distribution alignment~\cite{dann,dsn} or self-supervised learning~\cite{selfda1}. 

Multi-domain generalization (MDG) is an alternative framework where the goal is to learn domain-invariant features from multiple labelled source domains. Recent works~\cite{metadg,dgm} give a theoretical proof that domain-invariant features can be learned using MDG. The common goal of UDA and MDG is to learn domain-invariant features for downstream tasks (i.e. classification, segmentation and object detection) by training a model on data from multiple domains. Therefore, UDA and MDG both require access to data from multiple domains during training. For single domain generalization, M-ADA~\cite{sdg_adversarial2} and L2D\cite{sdg_adversarial1} propose an adversarial training framework for SDG learning. M-ADA proposes a meta-scheme method to find the adversarial perturbation, which calculates the gradient direction for each specific input and is extremely slow for real-life applications. While similar to RandConv~\cite{randconv} and GIN~\cite{gin}, M-ADA also proposes to synthesize new domain but with an adversarial training. However, M-ADA constrains the perturbation in latent space and cannot guarantee the semantic consistency for segmentation task, i.e object boundary and shape, which is especially crucial for medical image segmentation.

\paragraph{Data Augmentation}
Data Augmentation is a low-level data sampling technique, which gains a free performance improvement with use of human prior. To further improve the generalizability of models, AdaTransform~\cite{ada} proposes to augment the data by maximizing the entropy of the data. AdvBias~\cite{advbias} also apply the adversarial technique for learning the bias-field and deformation field respectively. Cutout~\cite{cutout} randomly samples masks and crops out part of images via the masks. Mixup~\cite{mixup} simply mixes the images and also forces the prediction of mixed images to be the interpolation of ground truth labels.
\section{Proposed Method}

In the single domain generalization setting, we aim at training a segmentation model $S$ on a single source domain $\mathcal{S}=\{x^i,y^i\}_{i=1}^N$ such that the learned model can generalize well to multiple target domains $\mathcal{T}_k=\{x^j_k,y^j_k\}_{j=1}^M$, where $k$ is the domain indicator. The target domains are not accessible during training.

We propose an adversarial training-based method to synthesize data from new domains $\mathcal{\hat{T}}_l=\{\hat{x}^i_l,y^i\}_{i=1}^N$ from the single source domain, where $l$ is the synthetic domain indicator. An assumption of our method is that the space of the collections of the unseen domains belongs to the collections of synthetic domains, i.e. $\cup_k\mathcal{T}_k \subset \cup_l \mathcal{\hat{T}}_l$. Specifically, the unseen domains can be generated by interpolating the synthetic domains. Our adversarial framework can enlarge the diversity the synthetic domains and guarantee sufficient coverage of the unseen domains. To ensure the synthetic images adhere to the semantics of the source images, we also apply a regularizer that promotes the mutual information between synthetic images and source images.

\begin{figure}[t]
\centering\includegraphics[width=12cm]{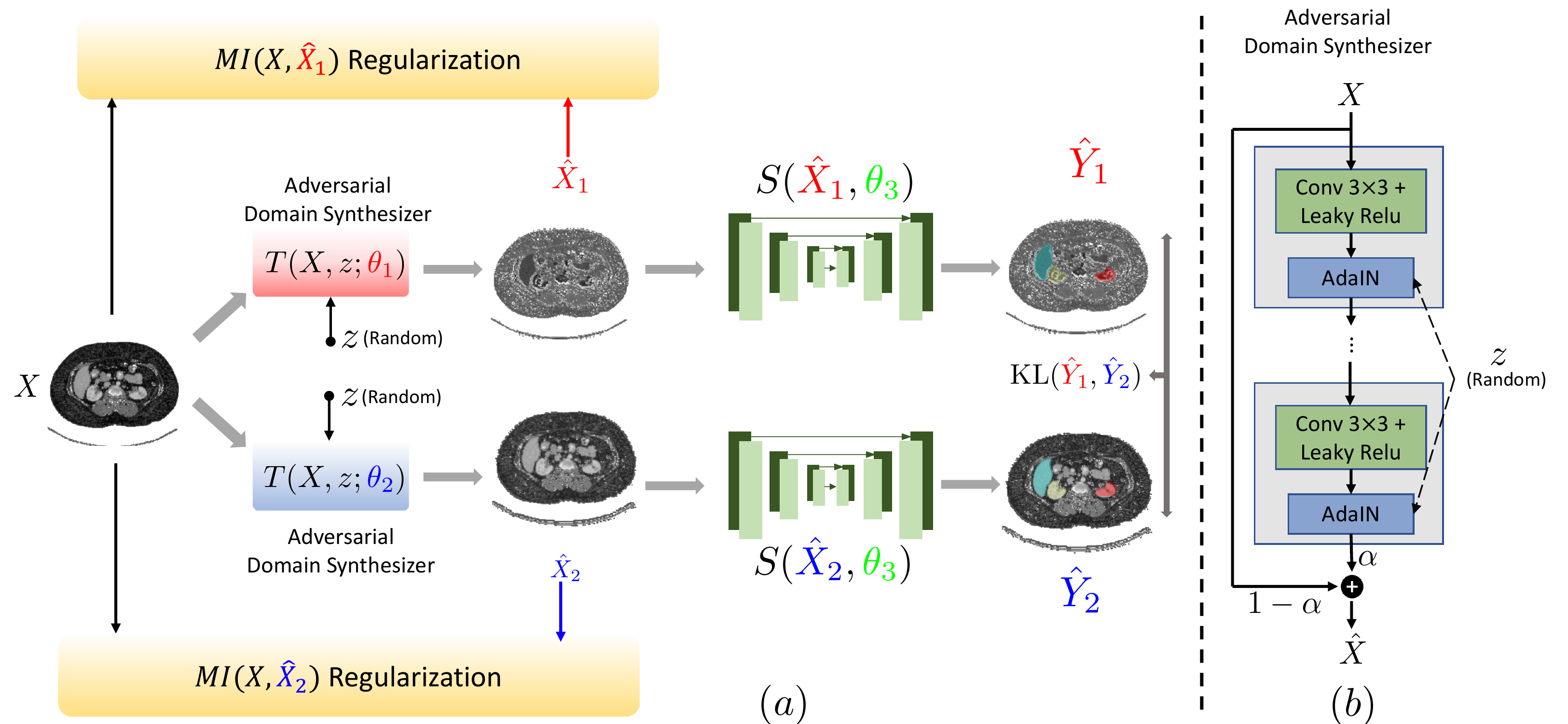}
  \caption{Schematic of the proposed method. (a) shows the completed model structure, which consists of our adversarial domain synthesizer (ADS), the mutual information regularization between input and synthetic image, the segmentation network and the KL consistency loss between two predictions. (b) is the detailed structure of our proposed ADS.}
  \label{schematic}
  \vspace{-3mm}
\end{figure}

Our method consists of four modules: two adversarial domain synthesizers paramaterized as ${T(\cdot;\color{red} \theta_1})$ and ${T(\cdot;\color{blue} \theta_2})$, a segmentation network parameterized as ${S(;\color{green} \theta_3})$ and the mutual information regularizer $\text{MI}(\cdot,\cdot;\theta_4)$. We train the models with a two-step min-max procedure that involves supervised, consistency, adversarial, and regularization loss terms. A schematic illustration of our proposed model is shown in Figure~\ref{schematic}.

\subsection{Adversarial Domain Synthesizer}\label{ADS method}
To maximize the effectiveness of the synthetic domains, we utilize adversarial training to learn an Adversarial Domain Synthesizer (ADS) $T(X,z;\theta)$, which takes as input a source image and random noise sample $z$ and outputs a synthetic image $\hat{X}$. We assume that $T$ only changes the texture of the source images and not their segmentation annotation. As a result, the segmentation network $S$ should generate the same segmentation mask for all $\hat{X}$ from a given $X$. For the adversarial training, $T$ competes with $S$ to generate the adversarial domain images. We construct the adversarial consistency via two differently parameterized $T$, and we can obtain two randomly synthesized images ${\color{red}\hat{X}_1}={T(X,z;\color{red} \theta_1}), {\color{blue}\hat{X}_2}={T(X,z;\color{blue} \theta_2})$. The purpose of random variable $z$ will be explained below. The adversarial competition can be formulated as follows:
\begin{equation}\label{adversaral}
\ell_{Con}({\color{red}\hat{X}_1},{\color{blue}\hat{X}_2}) = \text{KL}(S({\color{red}\hat{X}_1};{\color{green} \theta_3}) || S({\color{blue}\hat{X}_2};{\color{green} \theta_3})),
\end{equation}
where KL divergence is measured between the two softmax outputs. In Equation~\ref{adversaral}, the two synthesizers $T$ are trained to maximize $\ell_{Consistency}$ and $S$ is trained to minimize it.
\paragraph{ADS network architecture}
The goal of the ADS $T$ is to generate the synthetic domains with hardest perturbation for $S$. Meanwhile, the semantics of the images, such as the boundary and shape of the organs, should keep consistent with source images and the changes are limited to the texture of the tissues. To achieve the effect mentioned above, we proposed a modified version of Global Intensity Non-Linear Augmentation module (GIN)~\cite{gin}. GIN utilizes a shallow CNN structure with randomly initialized weights for the convolution filters plus a mixup with ratio of $\alpha \sim U(0,1)$ to generate multi-modalities from source domain. Unlike GIN, in our method, the synthesizer $T$ is learned adversarially, so we cannot initialize $T$ randomly to generalize multiple modalities. Thus, without randomness from the parameters of $T$, we can only interpolate the domains with the mixup parameter, which limits the variance of synthetic domains. To alleviate this issue, we incorporate the adaptive instance normalization (AdaIN) block~\cite{adain} to introduce randomness to the generated domain. The design of the proposed $T$ is shown in Figure~\ref{schematic} (b).
\subsection{MI Regularization}
\begin{figure}[t]
\vspace{-3mm}
\centering\includegraphics[width=10cm]{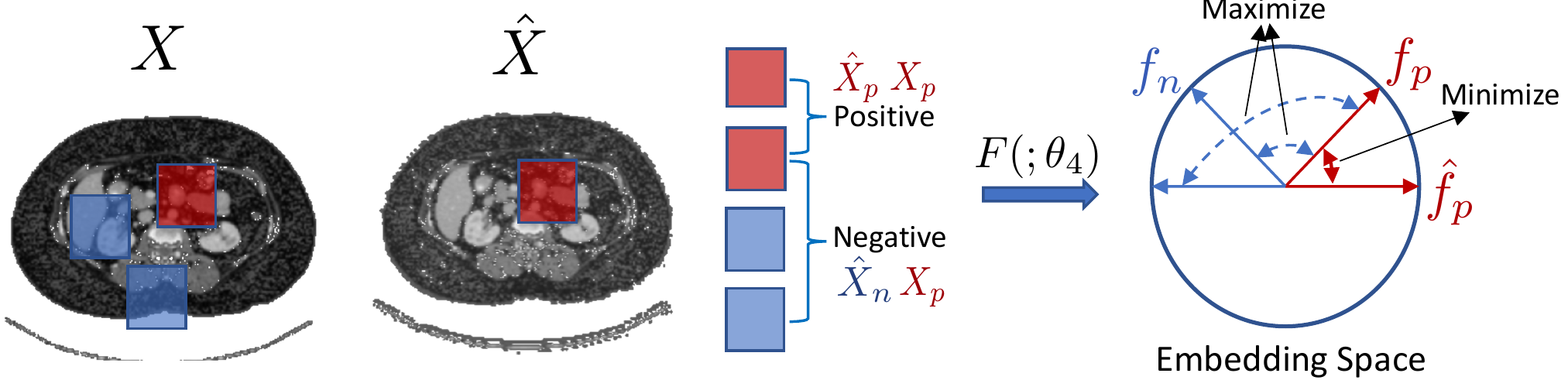}
  \caption{Mutual information maximization.}
  \label{MI model}
  \vspace{-3mm}
\end{figure}
Without proper regularization, $T$ can change semantics of the image arbitrarily. To ensure that the boundary and shape are shared underlying  semantics between the synthetic image and the source image, we can maximize the mutual information between two images to keep the semantics. Thus, we propose to add the mutual information maximization (MIM) constraint to the synthesized images as maximizing $\text{MI}(X, \hat{X})$, where $\theta_4$ is the parameters of the mutual information estimator. Exact calculation of mutual information is computationally prohibitive. Inspired by the success of contrastive learning in image-to-image translation~\cite{cut}, we use patch-based contrastive loss as a surrogate for MI. Specifically, we utilize constrastive learning on the image patch level to maximize the MI between the source images and the transformed images, which exactly satisfies our goal. Thus, we can adapt~\cite{cut} to our model and reformulate it as follows:
\begin{align}\label{MI equation}
    \ell_{MI}(X,\hat{X}) = \log &\frac{\exp(\hat{f}_p\cdot f_p/\tau)}{\exp(\hat{f}_p\cdot f_p/\tau)+\sum_{n\not= p}\exp(f_p\cdot f_n/\tau)}.
\end{align}
 In Equation~\ref{MI equation}, the $\ell_{MI}$ is maximized and the numerator is the inner product of two corresponding patch features with exponential re-scaling, the $\hat{f}_{p}$ and $f_{p}$ denote the features extracted by $F(\hat{X}_{p};\theta_4), F(X_{p}; \theta_4)$, where $\hat{X}_p, X_p$ represent the patches of the synthesized image and the source image at the corresponding location respectively and they are the positive pair for contrastive learning. For the negative pair, we can see the same operation of the $f_p$ and $f_q$ in the denominator. In Equation~\ref{MI equation}, we are trying to maximize the semantic correlation between the synthesized image and the input via pushing similarity of patches at the corresponding location and dissimilarity at different locations.
 \subsection{Model Optimization}
As mentioned above, the segmentation masks are kept consistent from the source domain to synthesized domains. Thus we also have the supervised loss as below:
\begin{equation}
    \ell_{Sup}(\hat{X},Y) = \text{KL}(S(\hat{X};{\color{green}\theta_3})||Y).
\end{equation}
Our two-step optimization for the proposed model can be summarized as:
\begin{align}\label{init obj}
    &{\text{update}~{\color{green}\theta_3}:}
    \min_{{\color{green}\theta_3}}\mathop{\mathbb{E}}_{X,Y\sim P_ {XY}}
    \ell_{Sup}({\color{red}\hat{X}_1},Y)+\ell_{Sup}({\color{blue}\hat{X}_2},Y) +\ell_{Cons}({\color{red}\hat{X}_1},{\color{blue}\hat{X}_2}) \nonumber\\
    &{\text{update}~{\color{red}\theta_1},{\color{blue}\theta_2},\theta_4:}
    \max_{{\color{red}\theta_1},{\color{blue}\theta_2},\theta_4}\mathop{\mathbb{E}}_{X\sim P_ {X}, z\sim P_z}\ell_{Con}({\color{red}\hat{X}_1},{\color{blue}\hat{X}_2}) + \ell_{MI}(X,{\color{red}\hat{X}_1}) + \ell_{MI}(X,{\color{blue}\hat{X}_2}) \nonumber.
\end{align}
\vspace{-3mm}
\section{Experiments}
\vspace{-3mm}
We evaluate our model on two experimental settings. For the first setting, we test the generalizability of the model on an unseen modality by using abdominal CT scans from~\cite{abdominal_mri} and MRI scans from~\cite{chaos} as the source domain and target domain respectively. For the second setting, we aim to test the generalizability of our model on unseen data distribution shift caused by different scanner machine settings. We obtain the data from six different organizations (RUNMC,BMC,HCRUDB,UCL,BIDMC,HK) and adopt the following cross-validation procedure: we select the data from one organization as the source domain for training and hold out the rest as the target domains for testing. Furthermore, we conduct a detailed ablation study to analyze the effectiveness of each component of the proposed method.
\vspace{-3mm}
\subsection{Training Configuration}
We use Efficient-b2~\cite{efficientnet} as the backbone for the segmentation network $S$ and modify it with UNet-style skip connection. The synthesizer network $T$ has 4 convolutional blocks, each block consisting of $3\times 3$ convolutional kernel with the channel size of 2, Leaky-ReLU and AdaIN. For the MIM model, we use the encoder part of the generator with a fully connected layer as the feature extractor $F$ from~\cite{cut}. We choose Adam optimizer~\cite{adam} for all of the models with initial learning rate $3\times 10^{-4}$ and $\beta=[0.5,0.999]$. The learning rate will linearly decay to zero at the end of the training. The total training epochs are 2,000. We compare our method to Cutout, AdvBias, RandConv and GIN~\cite{gin} under the same settings.
\vspace{-3mm}
\subsection{Data Prepossessing and Evaluation Metrics}
Before training and testing, we first resize all of the 3D volumes at the axial plane to the size of $192 \times 192$. For the CT modality, we clip the intensity into the range of $[-275, 125]$. For the MRI modality, the top $99.5\%$ of intensity values are cut out. We also apply all variations of predefined augmentations to the 3D volumes before we slice them into 2D images. These augmentations include random contrast via gamma transformation, Gaussian noise addition, affine transformation, 3D elastic transformation and intensity normalization of zero mean and unit variance. All of these predefined augmentations are performed with MONAI. More details of the data prepossessing and augmentation steps can be found in Appendix A.

To evaluate single source domain generalization, we do not access the target testing dataset during training. Thus, we split all source datasets into training and evaluation partitions with proportions $70\%$ and $30\%$, respectively. For the final testing on the target datasets, we pick the saved model checkpoint which has the best predictive performance on the source evaluation split. We use the Dice score (DS) for evaluating the performance of the different methods.
\begin{figure}[t]
\centering\includegraphics[width=12cm]{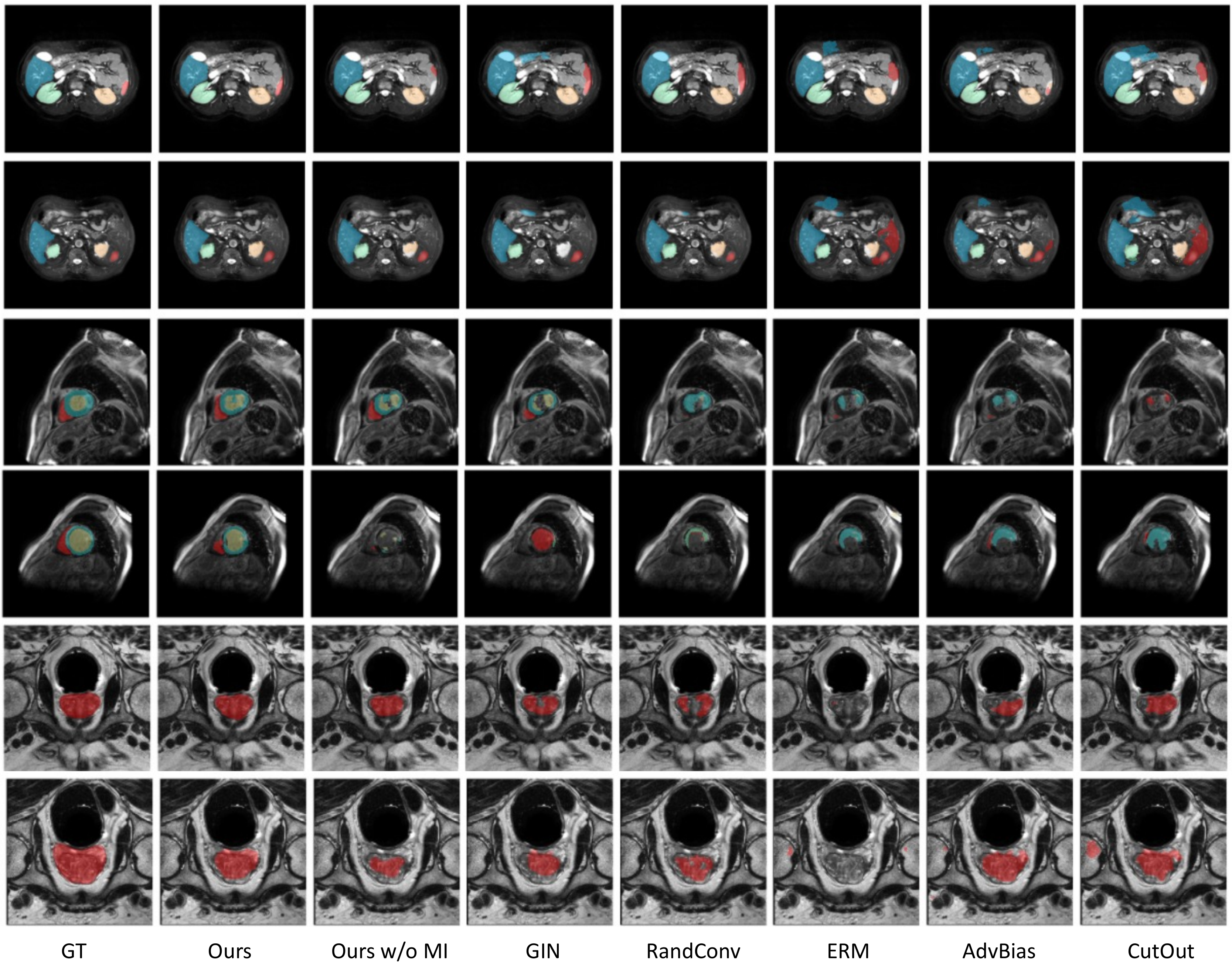}
  \caption{Visualization of results.}
  \label{seg visualization}
  \vspace{-3mm}
\end{figure}
\vspace{-3mm}
\subsection{Experimental Results and Empirical Analysis}
We summarize our results in Table~\ref{main results}. Across three different single domain generalization settings, our proposed method achieves the best performance compared with the baseline models. Notably, compared with~\cite{gin}, the overall performance on Cardiac bSSFP-LGE is worse than their reported results, although we have tried our best to replicate the baseline results under the suggested settings of~\cite{gin}. We also visualize the segmentation results in the Figure~\ref{seg visualization}. Our method succeeds in some difficult samples, where the partial spleen is missing or mis-classified in the Abdominal CT-MRI, or missing semantic structure in Cardiac bSSFP-LGE. We also observe the coarse segmentation mask in Prostate Cross-Centers for the baseline methods. The above quantitative and qualitative results show the effectiveness of our adversarial training framework. more visual results are shown in Appendix B. 
  \begin{table}[t]
   \caption{Results on three single source domain generalization settings.}\label{main results}
  \centering
  \resizebox{1.0\textwidth}{!}{
 \begin{tabular}{c|ccccc|cccc|cc}
  \toprule
  \multirow{2}{*}{Method}   & \multicolumn{5}{c|}{\textbf{Abdominal CT-MRI}}  & \multicolumn{4}{c|}{\textbf{Cardiac bSSFP-LGE}} & \textbf{Prostate Cross-center} \\
    & Liver & R-Kidney & L-Kidney & Spleen & Average & L-ventricle & Myocardium & R-ventricle & Average & Average \\
    \midrule
    ERM & 73.35 & 75.92 & 74.20 & \textbf{82.93} & 76.60 & 51.19 & 72.7 & 70.40 & 64.76 & 51.57\\
    Cutout & 76.57 & 80.06 & 77.96 & 78.90 & 78.37 & 65.13 & 77.58 & 70.44 & 71.05 & 58.79\\
    AdvBias & 74.97 & 86.16 & 78.39 & 72.45 & 77.99 & 60.10 & 77.52 & 72.29 & 69.97 & 60.47\\
    RandConv & 78.07 & 83.31 & 80.69 & 80.23 & 80.58 & 67.48 & 85.39 & 80.46 & 77.78 & 68.14\\
    GIN & 83.16 & 85.99 & 82.17 & 82.22 & 83.39 & 71.22 & 84.10 & \textbf{82.06} & 79.12 &  70.15\\
    Ours & \textbf{85.24} & \textbf{89.87} & \textbf{86.92} & 81.69 & \textbf{85.93} & \textbf{73.37} & \textbf{86.10} & 81.45& \textbf{80.31} &  \textbf{71.42}\\
  \bottomrule
 \end{tabular}}
 \vspace{-6mm}
\end{table}
\vspace{-3mm}
\subsection{Ablation Study}
\vspace{-3mm}
We conduct extra experiments to study the effectiveness of each component in our model. In Table~\ref{ablation perturbation}, we compare the performance results of our model without adversarial training, our model without the mutual information regularizer, and our completed model. This experiment is conducted in the Abdominal CT-MRI under the same setups as described above. The model without adversarial training still achieves better results than the model without mutual information regularization by a small gap. However, we can see that both adversarial training and the MI regularizer can coorporate with each other to achieve superior performance.
  \begin{table}[h]
  \vspace{-3mm}
  \caption{Ablation on Abdominal CT $\to$ MRI.}\label{ablation perturbation}
  \centering
  \resizebox{0.6\textwidth}{!}{
 \begin{tabular}{cccccc}
  \toprule
   \multicolumn{6}{c}{Abdominal CT-MRI} \\
    \midrule
     Method & Liver & R-kidney & L-kidney & spleen & average \\
    \midrule
    Ours w/o adversarial  & 84.59 & 85.52 & \textbf{87.79} & 79.21 & 84.28 \\
    Ours w/o MI & 83.23 & 85.62 & 83.71 & \textbf{82.79} & 83.59 \\
    Ours & \textbf{85.24} & \textbf{89.87} & 86.92 & 81.69 & \textbf{85.93} \\
  \bottomrule
 \end{tabular}
 }
 \vspace{-8mm}
\end{table}
\vspace{-3mm}
\section{Conclusion}
\vspace{-3mm}
We successfully trained a model on a single source domain and deploy it on unseen target domains for the purpose of \emph{single} domain generalization (SDG). Because of the limited training data and the unpredictable target domain shift, the generalizability of segmentation models is restricted, especially for medical images. Therefore, we propose a novel adversarial training framework to improve the generalizability of the SDG segmentation model. We synthesize the new domains via learning an adversarial domain synthesizer (ADS) in the proposed method. We assume that the unseen domains can be interpolated from synthetic domains, and the adversarial synthetic domains can guarantee sufficient coverage. To constrain the semantic consistency between synthetic images and the corresponding source images, we propose a mutual information regularizer, which can be estimated by patch-level contrastive learning. We evaluate our method for various organ segmentation for unseen modalities, scanning protocols, and scanner sites. The proposed method shows a consistent improvement over the baseline methods.
\vspace{-3mm}
\section{Acknowledge}
\vspace{-2mm}
This work was partially supported by NIH Award Number 1R01HL141813-01,
NSF 1839332 Tripod+X, SAP SE, and Pennsylvania Department of Health. We
are grateful for the computational resources provided by Pittsburgh SuperComputing grant number TG-ASC170024.
MG is supported by Australian Research Council Project DE210101624. KZ would like to acknowledge the support by the National Institutes of Health (NIH) under Contract R01HL159805, by the NSF-Convergence Accelerator Track-D award \#2134901, and by the United States Air Force under Contract No. FA8650-17-C7715.
%
%
%
\bibliographystyle{splncs04}
\bibliography{MICCAI2022-332}
\end{document}